\documentclass{article}

    \usepackage[preprint]{neurips_2020}

\usepackage[utf8]{inputenc} 
\usepackage[T1]{fontenc}    
\usepackage{hyperref}       
\usepackage{url}            
\usepackage{booktabs}       
\usepackage{amsfonts}       
\usepackage{nicefrac}       
\usepackage{microtype}      

\usepackage{soul}
\usepackage{amsmath}
\usepackage[pdftex]{graphicx}
\usepackage{caption}
\usepackage[dvipsnames]{xcolor}
\usepackage[normalem]{ulem}
\usepackage{csvsimple}

\newtheorem{theorem}{Theorem}[section]

\usepackage[normalem]{ulem}

\usepackage{array}
\newcolumntype{H}{>{\setbox0=\hbox\bgroup}c<{\egroup}@{}}

\title{Supervised Whole DAG Causal Discovery}

\author{%
  Hebi Li\\
  Dept. of Computer Science\\
  Iowa State University\\
  Ames, IA 50010 \\
  \texttt{hebi@iastate.edu}\\
  \And
  Qi Xiao\\
  Dept. of Electrical Engineering\\
  Iowa State University\\
  Ames, IA 50010 \\
  \texttt{qxiao@iastate.edu}\\
  \And
  Jin Tian \\
  Dept. of Computer Science\\
  Iowa State University\\
  Ames, IA 50010 \\
  \texttt{jtian@iastate.edu} \\
}

\begin{document}

\maketitle

\begin{abstract}

We propose to address the task of causal structure learning from data in a supervised manner. 
Existing work on learning causal directions by supervised learning is restricted to learning pairwise relation, and not well suited for whole DAG discovery. We propose a novel approach of modeling the whole DAG structure discovery as a supervised learning. 
To fit the problem in hand, we propose to use permutation equivariant models that align well with the problem domain. We evaluate the proposed approach extensively on synthetic graphs of size 10,20,50,100 and real data, and 
show promising results 
compared with a variety of previous approaches.


\end{abstract}

\section{Introduction}
\label{intro}

Causal structure discovery is an important topic for artificial intelligence
research. We will focus on the task of learning from data a Directed Acyclic Graph (DAG) where an edge in the DAG indicates the direct causal relationship between the parent node and the child node. There have been many methods proposed to recover causal structures among
variables from observational data and/or interventional data.

There are several different kinds of approaches for causal discovery. Traditional
constraint-based algorithms such as PC~\cite{2009-Book-Pearl-Causality} and FCI~\cite{2000-Book-Spirtes-Causation} rely on conditional independence tests to
infer graphical conditions, and construct causal graphs without violating the
constraints. Score-based methods like GES~\cite{2002-JMLR-Chickering-Learning, 2017-Journal-Ramsey-Million} and CAM~\cite{2014-Journal-Buhlmann-CAM} use a pre-defined score-function, such as AIC,
BIC, BDe scores, and search in the space of all valid DAGs for the highest
possible score, via search algorithms such as Hill Climbing.

A recent break through, NOTEARS~\cite{2018-NIPS-Zheng-Dags}, models the acyclic constraint as an equivalent continuous
equality constraint, and solves the causal discovery entirely as a continuous
optimization problem via augmented Lagrangian.
Follow-up works along this line design different search strategies and non-linear extensions: DAG-GNN~\cite{2019-ICML-Yu-Dag} extends the linear model to non-linear via generative models; RL-BIC2~\cite{2020-ICLROral-Zhu-Causal} uses reinforcement learning as the search strategy; and Grad-DAG~\cite{2020-ICLRPoster-Lachapelle-Gradient} also extends the model to non-linear using neural networks.

Another line of research models pair-wise causal discovery problem as a supervised learning, to infer the causal direction
between a pair. Work along this line includes Randomized Causation Coefficient (RCC)~\cite{2015-ICML-Lopez-Paz-Towards, 2015-JMLR-Lopez-Paz-Randomized} and Manifold Regularized Causal Learning (MRCL)~\cite{2019-JMLR-Hill-Causal}. However, both works are pairwise methods, and do not work well in discovering global DAG structure.



We propose a novel approach of modeling the whole DAG structure discovery as a supervised learning.  In particular, we learn a predictive model that takes a featurization of the input data, and directly predicts the whole DAG structure. The model is trained in a supervised manner, where the supervised data is generated from
randomly sampled DAGs. When performing causal discovery on testing data, the
resulting DAG is obtained efficiently by a single forward inference pass of
the model, instead of searching or optimization.
We focus on the linear causal models in this paper.





We choose to model the learning problem using a neural network.
The problem, however, is quite challenging to fit by na\"ive models. 
Experimental results show that 
na\"ive Fully-Connected (FC) networks and Convolutional Neural Networks (CNNs) do not work well due to misalignment between the characteristics of the model and the problem domain. Our key observation is that the problem at hand is permutation equivariant w.r.t variable orders. That is, when the input variable order is permuted, the output adjacency matrix should change accordingly with the same permutation applied. Motivated by this observation, we instead utilize the recently proposed permutation Equivariant models~\cite{2017-NIPS-Zaheer-Deep, 2018-ICML-Hartford-Deep}, and show significant improvement over baseline FC models and CNN models. We call this new method by \emph{DAG-EQ}, shortcut for \emph{DAG Structure Discovery by Equivariant Models}.
Similarly, we refer to the baselines that use FC and CNN models as \emph{DAG-FC} and \emph{DAG-CNN}, respectively.





We conduct extensive evaluation of proposed DAG-EQ on different types of graphs of sizes 10, 20, 50, 100, and compare with a variety of methods, including traditional constraint-based method (PC), score-based methods (GES, CAM), continuous optimization methods (NOTEARS, DAG-GNN, RL-BIC2), and  pairwise supervised methods (RCC) 
modified to learn the DAG structure.
We show our method performs much better than PC, GES, CAM, RCC, and is comparable with state-of-the-art NOTEARS-based approaches.
We also test DAG-EQ on real data of protein signaling network~\cite{2005-Science-Sachs-Causal}, and show comparable results with state-of-the-art methods. 
Notably, DAG-EQ pre-trains a predictive model once with randomly generated examples. Subsequently the model performs causal discovery on all new tasks very efficiently - no search or optimization needed. Experimental results show that the pre-trained models perform well across graphs of different types, sizes, or causal strengths.

In summary, we make the following contributions:

\begin{enumerate}
\item 
We propose a novel formulation of whole DAG structure discovery  as a supervised learning problem. This is the first method of such kind to our best knowledge.
\item We propose to use equivariant models
to fit the statistical learning problem. 
We develop DAG-EQ, a DAG structure learning algorithm based on equivariant deep neural networks that capture the intrinsic characteristics of the data.
\item We evaluate DAG-EQ extensively on synthetic datasets and real data, and show promising results 
compared with a variety of previous approaches.

\end{enumerate}

\section{Background: Functional Causal Model}

Functional causal model~\cite{2009-Book-Pearl-Causality} has been the
standard causal framework in the literature. For a set of random
variables $\{X_1, \dots, X_d\}$, where $d$ is the number of variables, a functional
causal model consists of a set of equations of the form
\begin{equation}
X_i = f_i(pa_i, \epsilon_i), \quad\quad i=1, \dots, d
\end{equation}
where $pa_i$ (connoting parents) stands for the set of variables
that are immediate causes of $X_i$, and $\epsilon_i$ represents
error term due to unobserved factors.
The linear Structural Equation Models (linear SEMs) is a linear instance of
functional causal model, and is given by
\begin{equation}
  X_j = \sum_{i \ne j} c_{ji} X_i + \epsilon_j, \quad\quad j= 1,\dots,d
\end{equation}
Here, each equation characterizes the direct causal influence of $X_i$
on $X_j$, quantified by the path coefficient $c_{ji}$. 
We consider the
recursive models in which we assume the coefficients $c_{ji} = 0$ for
$i<j$. Thus, the set of variables ordering is directed and acyclic. We
denote the covariances between observed variables as $\sigma_{ij} =
Cov(X_i, X_j)$, and covariances between error terms as $\phi_{ij} = Cov(\epsilon_i,
\epsilon_j)$. We define matrix $\Sigma=[\sigma_{ij}]$,
$\Phi = [\phi_{ij}]$, and $C=[c_{ij}]$ accordingly.

The model structure can be represented by a directed acyclic graph
(DAG) G, called the causal diagram, as follows: the nodes of G are
variables $X_1, \dots, X_d$; there is a directed edge from $X_i$ to
$X_j$ if $c_{ji \ne 0}$; there is a bidirected edge between $X_i$ and
$X_j$ if error terms $\epsilon_i$ and $\epsilon_j$ have non-zero
correlation ($\phi_{ij} \ne 0$). The covariance matrix $\Sigma$ is
given by
\begin{equation}\label{sigma}
  \Sigma = (I-C)^{-1} \Phi (I-C)^{T^{-1}}
\end{equation}

\section{Approach}
In this section, we formulate the DAG structure discovery as a supervised learning problem. 
Since the common learning models cannot take a distribution as input, we first discuss how we featurize the distribution into a feature vector suitable as model input. Then, we discuss na\"ive baseline models and the permutation equivariant model that we use to fit the problem. Finally, we discuss the training objective and inference procedure.

\subsection{Featurization}

The structure learning task takes a sample of data points $X=\{X_i\}^N$ as inputs, and outputs adjacency matrix of all the variables $\{X_1,\ldots,X_d\}$. However, the common learning models cannot take a distribution as input. Thus, we need to first embed the input data distribution as a feature vector,
and train a model to predict the adjacency matrix for the feature vector.


Motivated by the covariance matrix equation of SEMs in Eq. (\ref{sigma}), a natural choice is to use covariance matrix as the featurization, and the learning problem is essentially to learn a model to simulate Eq. (\ref{sigma}). However, the scale of covariance matrix is sensitive to the data distribution, 
such that models learned on training data sets do not work/transfer well to datasets with different coefficients scales.
We thus use the normalized version of the covariance matrix, the \textit{Pearson correlation coefficients}, also called the \textit{correlation matrix}, defined as:

\begin{equation}
    \rho_{X,Y} = \frac{cov(X,Y)}{\sigma_X \sigma_Y}
\end{equation}

Pearson correlation coefficients are normalized into the range of $[0,1]$, thus transfer much better than covariance matrix when training and testing distributions differ. 

In summary, given the input data distribution over $X$, we compute correlation matrix $\rho_{X,X}$ which is a $d \times d$ matrix, as the input feature vector.


\subsection{Baseline Models}

Recall that our goal is to learn a mapping from correlation matrix of size $d \times d$ to the adjacency matrix of size $d \times d$  where $d$ is the number of variables. In particular, the learning model takes as input the correlation matrix $\rho_{X, X}$ of the input data distribution $X$, and
outputs the edge probability matrix $\widehat{Y}$.
\begin{equation}
\widehat{Y} = f_\theta(\rho_{X,X})
\end{equation}
where $f_\theta$ is the model parameterized by $\theta$.

In this paper, we consider using neural networks for better model expressiveness. We consider two baselines, the Fully-Connected (FC) Multi-Layer Perceptrons (MLPs) and CNN models.

The FC models first reshape the input matrix to one dimension vector of size $d^2$. The vector is fed into multiple fully-connected layers with a ReLU non-linear activation after each layer. The output of the last layer is of size $d^2$, and reshaped to $d \times d$ matrix representing the probability of the existence of edges in the adjacency matrix. These probabilities are activated by sigmoid functions to ensure the probabilities are in $[0,1]$. 



For the CNN baseline, we consider
a ``flat'' CNN where all convolutional layers use $3\times 3$ kernels with padding size 1, so that the feature maps of all layers are of the same size.
The output is used as final adjacency matrix after a sigmoid activation.

\subsection{Permutation Equivariant Models}
As we shall show in the evaluation, the two baseline models cannot fit the problem well, especially for large graphs. They tend to over-fit the training data. The FC model does not enforce any kinds of regularization; the CNN models enforce local relations, but in our problem domain, 
the relations between variables are  long range.
Thus there needs some strong form of regularization better aligned with the problem domain.


The key observation is that, 
when the input variable order is permuted, the output adjacency matrix should change accordingly with the same permutation applied. Formally, for any
permutation matrix $P$ and input correlation matrix $\rho_{X,X}$, the model $f_\theta$ should
satisfy the following property:
\begin{equation}
f_\theta(P \rho_{X,X} P^T) = P f_\theta(\rho_{X,X}) P^T
\end{equation}

This is known as permutation
equivariant~\cite{2016-ICML-Cohen-Group,2017-NIPS-Zaheer-Deep}. Formally,

\begin{theorem}\cite{2017-NIPS-Zaheer-Deep}
Let $f$ be a function of the form of $f(x) = \sigma(Wx)$, i.e. a 1-layer neural network with weight $W$ and 0-bias, where $x \in \mathbb{R}^M$, $W \in \mathbb{R}^{M\times M}$ i.e. $f: \mathbb{R}^M \rightarrow \mathbb{R}^M$. The function $f$ is permutation equivariant iff $W$ has the form

\begin{equation}
    W = \lambda \mathbf{I} + \gamma (\mathbf{11}^T) \quad\quad \lambda, \gamma \in \mathbb{R} \quad \mathbf{1}=[1,\ldots,1]^T \in \mathbb{R}^M \quad \mathbf{I} \in \mathbb{R}^{M \times M} \text{is the identity matrix}
\end{equation}
\end{theorem}

In our case, the
input and output are 2-dimentional matrix, thus we use a generalization called
\textit{matrix permutation equivariant model}~\cite{2018-ICML-Hartford-Deep}, formally

\begin{theorem}\cite{2018-ICML-Hartford-Deep}\label{theorem:matrix}
Let $f=\sigma(W vec(X))$. $f$ is an \textit{exchangeable matrix layer} iff the elements of the parameter matrix $W$ are tied together such that the resulting fully connected layer simplifies to
\begin{equation}
\small
Y = \sigma(w_1 X + w_2 \mathbf{11}^T X +  w_3 X \mathbf{11^T} + w_4 \mathbf{11}^T X \mathbf{11}^T + b)
\end{equation}
where $\mathbf{1} = [1,\ldots,1]^T$ and $w_1,\ldots,w_4,b \in \mathbb{R}$
\end{theorem}

Intuitively, an equivariant kernel has 5 components, $w_1, w_2, w_3, w_4$ and $b$. In particular, $w_1$ is a constant weight applied to all entries of the input matrix. $w_2$ and $w_3$ are weights applied to sums of rows and columns, respectively. $w_4$ is the weight to the sum of the whole input matrix, and $b$ is bias. All of these 5 components are equivariant to permutations of rows and columns, thus the whole equivariant operator is equivariant to matrix permutation.



Such layers can be stacked to form a deep network. The capacity of the equivariant layers can be tuned by changing the number of channels. In particular, each layer can have
multiple such equivariant kernels, producing multiple channel outputs. The next equivariant layer is applied individually to each of the channel, and the results are summed to form the output. This procedure is in the same way as the channels of Convolutional layers in CNNs.


One additional feature for such stacked equivariant layers is that it natually accepts varying sizes input. Thus, it is possible to ensemble train the model on graphs of different sizes and types, and it is also possible to apply the trained models on varying and unseen graphs. The transferability and ensemble training results are presented in Evaluation section.


\subsection{Training and Inference}

As described above, we apply sigmoid activation to 
the model outputs,
so that the output probability matrix $\widehat{Y} \in [0,1]^{d \times d}$. We compute binary cross-entropy of $\widehat{Y}$ and true binary labels $Y \in \{0,1\}^{d \times d}$ as the loss function:

\begin{equation}
    L = -\frac{1}{N} \sum_{n=1}^N \sum_{j=1}^{d} \sum_{i=1}^{d}\big[ Y_{i,j}^{n} \cdot \log(\widehat{Y}_{i,j}^{n}) + (1-Y_{i,j}^{n}) \cdot \log(1-\widehat{Y}_{i,j}^{n}) \big],
\end{equation}
where $n$ is the sample index, and $i, j$ are the row and column index respectively.
The optimization is done via stochastic gradient descent using standard optimizer, in particular, we used the Adam~\cite{kingma2014adam} optimizer in our evaluation.



During the inference (the structure discovery), given a distribution of data points $\{X_i\}^N$, we perform the following steps to obtain the DAG: (1) compute featurization of the input data, in particular, the Pearson correlation matrix $\rho_{X,X}$, (2) perform model inference, $\widehat{Y} = f_\theta(\rho_{X,X})$, (3) 
construct DAG:
  we recursively add the edge with the highest probability (larger than $0.5$) unless it introduces a cycle.

\section{Evaluation}



Our model is implemented in the Julia programming language, and Flux~\cite{Flux.jl-2018} neural network library.  The source code 
and pre-trained models can be found at \url{https://github.com/lihebi/DAG-EQ}. For the methods in comparison, we obtain the python or R implementations from the authors.
All experiments are done in a desktop with Ryzen 3600 6-core 12-thread processor and a Nvidia RTX2060-Super graphical card.

\subsection{Evaluation Setup}
\paragraph{Evaluation metrics}
We report precision and recall of discovered edges, as
\begin{equation}
  \begin{aligned}
    \text{precision} & = & \frac{|\widehat{E} \cap E|}{|\widehat{E}|}\quad
    \text{recall} & = & \frac{|\widehat{E} \cap E|}{|E|}
  \end{aligned}
\end{equation}
where $\widehat{E}$ is the set of predicted edges, $E$ is the set of true edges. We also report Structural Hamming Distance (SHD),
which is the smallest number of edge additions,
deletions, and reversals to convert the estimated graph into the true DAG. Since PC and GES may output undirected edges, we treat undirected edges as bidirectional to calculate the metrics.


\paragraph{Synthetic data generation}

Following the literature, we use Scale-free (SF) and Erdos-Renyi (ER) graphs because they have been shown to be close to real
causal graphs. The number of edges of SF and ER graphs are the same as the number of nodes. For each setting with $d \in \{10, 20, 50, 100\}$ number of variables, we sample 1000   random
DAGs. 
We use $80\%$ as training graphs, and $20\%$ as testing graph, so that the model never sees the same graph in training and testing data. 

For each generated DAG, we sample 1000 data points according to a linear
causal model with additive Gaussian of $\mathcal{N}(0,1)$ or non-Gaussian noise described separately later. The linear causal model is generated by generating coefficients matrix $C_{ij}$ according to the sampled DAG, in the form of:
\begin{equation}
  \small
  \begin{split}
    C_{ij}=
    \begin{cases}
      \text{uniformly from} [-0.5-k,-0.5]\cup[0.5,0.5+k] & \text{ if } E_{i \rightarrow j} = \text{true}\\
      0 & \text{otherwise}
    \end{cases}
  \end{split}
\end{equation}
where $k$ is a hyper-parameter that controls the scale of the weight matrix. Finally, the true label $Y_{ij}$ is the binary adjacency matrix where 
$Y_{ij}=0$ if edge does not present, and $Y_{ij}=1$ otherwise.

Our data generating scheme will generate identifiable causal graphs: the linear SEMs with equal variance Gaussian noise are shown to be identifiable~\cite{2014-JMLR-Peters-Causal}; linear SEMs with non-Gaussian noise are also identfiable~\cite{2006-JMLR-Shimizu-Linear}.


\paragraph{Model details}
We refer to our model with equivariant layers as DAG-EQ, and denote those with baseline FC and CNN models as DAG-FC and DAG-CNN, respectively. The DAG-EQ model consists of 6 equivariant layers of hidden layer of size 300 (channels). There's a leaky-ReLU activation after each layer, following the recommendation in the literature~\cite{2018-ICML-Hartford-Deep}. The final output has the sigmoid activation to make sure the output is in the range of $[0,1]$.
The DAG-FC baseline model consists of 6 fully connected layers with hidden layer size 1024, and a ReLU non-linear activation after each layer. The DAG-CNN consists of 6 CNN layers, all using $3\times 3$ kernels with padding size 1. 
There's a ReLU non-linear activation and a batch normalization layer after each convolutional layer.

\paragraph{Existing methods in comparison}
We compare our model with pairwise supervised model, the Randomized Causation Coefficient (RCC)~\cite{2015-ICML-Lopez-Paz-Towards, 2015-JMLR-Lopez-Paz-Randomized}. However, since it only outputs the direction of the edge of either A to B or B to A, we modify the model to predict the existence of an edge from A to B. The existence of the reverse edge (B to A) is obtained by feeding the model with B and A in reverse order. 
During the inference, to predict the whole DAG, we apply the model to all pairs of variables. The original RCC used random forest classifier, denoted as RCC-RF. We also implemented a fully-connected neural network classifier, denoted as RCC-NN.

We also compare our methods with various types of structure learning approaches in the literature.  For traditional methods, we compare with constraint-based methods PC algorithm~\cite{2009-Book-Pearl-Causality}, score-based method Greedy Equivalence Search (GES)~\cite{2002-JMLR-Chickering-Learning,2017-Journal-Ramsey-Million} and Causal Additive models (CAM)~\cite{2014-Journal-Buhlmann-CAM}.
Continuous optimization approaches in the line of NOTEARS are state-of-the-art in terms of learning accuracy. In particular, we consider the original NOTEARS~\cite{2018-NIPS-Zheng-Dags}, and two most recent follow-up works, the DAG-GNN~\cite{2019-ICML-Yu-Dag} that uses a generative model with graph neural network, and RL-BIC2~\cite{2020-ICLROral-Zhu-Causal} that uses reinforcement learning to do the graph search.

\subsection{Comparison with the Literature}
We first present the structure discovery results on SF graphs of sizes 10, 20, 50, and 100, and compare our DAG-EQ with baseline DAG-FC, DAG-CNN and existing approaches. Results shown in Table \ref{tbl:literature}.


\begin{table}[ht]
\footnotesize
\centering
\caption{Comparison with literature, shown scale-free (SF) graph of size d=10,20,50,100\label{tbl:literature}}
\begin{tabular}{l|rrrrH||rrrrH}
\toprule
model   & d & prec  & recall & shd  & time (s) & d & prec  & recall & shd  & time (s)   \\
\midrule
DAG-EQ & 10 & 93.0 & 94.7   & 1.1  & 0.1 & 20& 92.3 & 89.2   & 3.5  & 0.1 \\
DAG-FC  &10   & 81.7 & 79.8   & 3.4 & -      &20& 49.0 & 41.6   & 19.3 & - \\
DAG-CNN &10  & 88.7 & 87.7   & 2.1 & -  &20& 82.0 & 78.4   & 7.4 & - \\
\midrule
RCC-RF &10& 17.0  & 96.7   & 44.4 & 3.26   &20& 9.4   & 97.9   & 192.6  & 12.39  \\
RCC-NN  &10& 18.6  & 68.9   & 33.2 & 3.15   &20& 11.9  & 75.3   & 122.5  & 12.18  \\
\midrule
PC   &10   & 26.9  & 35.6   & 14.3 & 1.69  &20& 33.9  & 47.9   & 26.9   & 1.84  \\
GES &10     & 19.5  & 30.0   & 15.6 & 0.65   &20& 19.1  & 26.3   & 34.3   & 0.75   \\
CAM    &10 & 7.5   & 25.6   & 35.5 & 5.24   &20& 6.6   & 30.5   & 94.3   & 16.62  \\
\midrule
NOTEARS &10& 100.0 & 100.0  & 0.0  & 1.04 &20& 92.8  & 94.7   & 2.5    & 14.92  \\
DAG-GNN &10& 92.3  & 88.9   & 1.7  & 65.93  &20& 84.8  & 92.1   & 5.0    & 70.00  \\
RL-BIC2  &10& 23.3  & 50.0   & 15.5 & 230.43  &20& 12.5  & 6.6    & 19.25  & 1099.77 \\
\midrule\midrule
DAG-EQ    &50  & 91.1 & 67.0   & 19.4 & 0.2 &100 &82.3 & 58.4   & 53.7 & 1.0  \\
DAG-CNN &50  & 50.0 & 43.4   & 49.0 & -   &100& 51.3 & 28.4   & 97.6  & -\\
\midrule
RCC-RF &50& 3.5   & 78.8   & 1113.9 & 76.14  &100& 2.4   & 69.5   & 2800.8 & 356.08  \\
RCC-NN  &50& 5.6   & 63.7   & 554.1  & 76.30   &100& 3.1   & 67.1   & 2130.8 & 355.02 \\
\midrule
PC     &50 & 33.8  & 43.3   & 68.2   & 2.34   &100& 34.8  & 45.5   & 138.0  & 66.49     \\
GES     &50& 10.7  & 15.9   & 104.6  & 0.81    &100& 6.9   & 10.9   & 233.2  & 1.37  \\
CAM     &50& 7.6   & 36.7   & 249.8  & 50.77    &100& 7.2   & 34.3   & 509.0  & 137.81 \\
\midrule
NOTEARS &50& 94.6  & 97.3   & 4.2    & 104.40  &100& 70.6  & 89.5   & 49.8   & 3286.48 \\
DAG-GNN &50& 82.1  & 91.2   & 16.0   & 75.49  &100& 79.3  & 87.7   & 37.8   & 79.10   \\

\bottomrule
\end{tabular}
\end{table}

We first compare DAG-EQ with baselines. DAG-EQ models perform much better compared with baseline DAG-FC and DAG-CNN. DAG-FC models can fit only for small graph of size 10, but struggle to fit for graphs of 20 or more. CNN models perform much better than FC models, however, it performs significantly worse than EQ models. This demonstrates that the characteristic of CNN does not align well with the problem domain. Instead, EQ models fit well on graph of sizes up to 100.

Comparing DAG-EQ with pairwise supervised model, RCC, we find that both RCC-RF and RCC-NN are not well suited for predicting the whole DAG. This suggests that pairwise model is insufficient to perform whole DAG discovery. A potential explanation is that they can't distinguish between direct and indirect effects, therefore they will add extra edges, causing low precision.

Comparing with traditional constraint-based and score-based methods, we find that PC, GES, and CAM do not perform well, especially when the graphs are large. This is consistent with the results reported by NOTEARS line of work~\cite{2018-NIPS-Zheng-Dags,2019-ICML-Yu-Dag,2020-ICLROral-Zhu-Causal,2020-ICLRPoster-Lachapelle-Gradient}. Our model can infer much more accurate graphs.

Finally, comparing with state-of-the-art NOTEARS line of research.
our model under-performs
depending on the size of the graphs.
For example, for a large graph d=100, our EQ model achieves SHD=53.7, while NOTEARS and DAG-GNN achieve SHD=49.8 and SHD=37.8 respectively. 

\subsection{Transferability and Ensemble Training}

It is desired to have one trained model to work for graphs of different
sizes and types.
We approach this goal in two
ways. First, we study the \textit{direct} transferability
across different noise models, graphs of different types and sizes, and different causal strengths.
Second, we ensemble train the model with graphs of varying sizes and kinds, and test the trained model on other unseen graphs.

Firstly, we apply DAG-EQ model trained on Gaussian noise model with $\mathcal{N}(0,1)$ and 3 other non-Gaussian noises, including Exponential with $\lambda=1$, 
Gumbel with $\mu=0,\beta=1$ and Poisson with $\lambda=1$. As shown in Table \ref{tbl:noise}, the model works very well on all different noise models.


\begin{table}[ht]
\centering
\small
\caption{Transfer between different noise models.\label{tbl:noise}}
\begin{tabular}{l|c|c|rrrr|rrrr}
\toprule
model   & train\_noise & test\_noise & d & prec & recall & shd & d & prec & recall & shd \\
\midrule
DAG-EQ            & Gaussian     & Gaussian & 10     & 87.2 & 96.9   & 1.6 & 20 & 86.4 & 93.9   & 4.0 \\
DAG-EQ      & Gaussian     & Exp    & 10           & 92.1 & 94.8   & 1.2 & 20 & 92.4 & 87.5   & 3.7\\
DAG-EQ          & Gaussian     & Gumbel  & 10         & 92.1 & 94.4   & 1.2 & 20 & 92.4 & 87.6   & 3.7\\
DAG-EQ          & Gaussian     & Poisson & 10      & 92.3 & 94.9   & 1.2  & 20 & 92.4 & 87.4   & 3.8\\
\bottomrule
\end{tabular}
\end{table}

We then transfer model to different causal strengths, defined by the weights of the causal coefficients, generated uniformly from the range $[-0.5-k,-0.5] \cup [0.5, 0.5+k]$. In Table \ref{tbl:scale}, the DAG-EQ model is trained on $k=1$, and tested on $k=1,2,4$. 
The transferability is more limited compared with that of different noise models. This reveals that the training and testing distribution didn't align with each other very well, and suggests two interesting research directions in the future. On the one hand, there requires a more flexible featurization that is trainable and adapts to the training and testing datasets. On the other hand, it is interesting to find a model that is invariant to the causal strength.


\begin{table}[ht]
\centering
\small
\caption{Transfer between different causal strengths.\label{tbl:scale}}
\begin{tabular}{l|rr|rrrr|rrrr}
\toprule
model & train/k & test/k  & d& prec & recall & shd & d & prec & recall & shd  \\
\midrule
DAG-EQ    &1   & 1 & 10           & 87.2 & 96.9   & 1.6 & 20 & 86.4 & 93.9   & 4.0  \\
DAG-EQ  &1      & 2  & 10         & 79.4 & 91.2   & 2.9 & 20  & 75.2 & 80.9   & 8.7 \\
DAG-EQ    &1     & 4    & 10           & 64.7 & 72.3   & 6.1  & 20   & 57.9 & 61.5   & 15.8 \\
\bottomrule
\end{tabular}
\end{table}

We also test the transfer performance across different types of graphs.
In particular, the DAG-EQ models are trained on ER graphs and tested on SF graphs, and vice versa. We show two results in Table \ref{tbl:ersf}. First, the the model transfers well across different types of graphs. For example, $ER\rightarrow SF$ performance is comparable to native performance of $SF \rightarrow SF$, and $SF\rightarrow ER$ is comparable to $ER \rightarrow ER$. Second, ER graphs seem to be harder than SF graphs for DAG-EQ model. 

\begin{table}[ht]
\centering
\small
\caption{Transferring between different types of graphs.\label{tbl:ersf}}
\begin{tabular}{l|c|c|rrrr|rrrr}
\toprule
model & train\_gtype & test\_gtype & d & prec & recall & shd & d & prec & recall & shd  \\
\midrule
DAG-EQ      & SF           & SF      & 10      & 87.2 & 96.9   & 1.6   & 20& 86.4 & 93.9   & 4.0\\
DAG-EQ      & ER           & ER       & 10     & 72.5 & 84.1   & 4.5   & 20   & 72.1 & 80.3   & 9.9\\
\midrule
DAG-EQ     & ER           & SF       & 10      & 82.0 & 95.4   & 2.3  & 20      & 80.9 & 84.7   & 6.7  \\

DAG-EQ     & SF           & ER    & 10         & 69.0 & 70.9   & 5.8  & 20    & 72.7 & 62.1   & 11.9 \\

\bottomrule
\end{tabular}
\end{table}

One potential advantage of DAG-EQ is that it can be trained once on relatively small graphs, and be applied to very large graphs where most existing methods are hard to scale. In this experiment, we apply the DAG-EQ trained using d=100 on very large graphs of sizes 200, 300, 400. We compare the results with GES, because it runs efficiently on graphs of this scale. Results are shown in Table \ref{tbl:ultra-large}. When the size increases, DAG-EQ can still maintain relatively good precision and recall. In particular, the recall remains 73\% for even d=400 graphs. On the other hand, precision degrades much more, due to increased number of positive predictions. In comparison, GES had much lower precision and recall. On d=300 and 400, GES had lower SHD, but much lower precision and recall. This is because GES predicted much smaller number of edges compared to DAG-EQ, and most of the predictions are false-positive. E.g. for d=300, GES predicted 742 edges, while only 19 of them are true-positives. DAG-EQ predicted 1211 edges but captured 232 true-positives. Overall, this results show that DAG-EQ have good transferability, and perform well on very large graphs where only a few methods can scale to.

\begin{table}[]
    \centering
    \footnotesize
    \caption{\small Transferring to very large graphs. Trained on d=100. \texttt{Pred}: predicted edges, \texttt{TP}: true positive.
    \label{tbl:ultra-large}}
    \begin{tabular}{l|rrrrrr||rrrrrr}
    \toprule
model   & d & Pred & TP & prec  & recall & shd  & d& Pred & TP & prec  & recall & shd \\
\midrule
DAG-EQ &100 & 70 & 58 &82.3 & 58.4   & 53.7 & 200 & 404 & 152 & 37.5    & 75.5    & 302.17   \\
GES  &100 & 169 & 13 & 7.6  & 13.13   & 242  & 200 & 431 & 19   & 4.4 & 9.5 & 592
\\
\midrule
\midrule
DAG-EQ & 300& 1211 & 232 & 19.1   & 75.5   & 1058.7  & 400 & 2984 & 330 & 11.1 & 73.5 & 2796.3\\
GES & 300 & 742 & 19 & 3.7 & 9.2 & 980.8 &    400& 1148& 28&  4.1& 11.6& 1443.2\\
\bottomrule
    \end{tabular}
\end{table}


Finally, we show that, the model can be ensemble-trained using graphs of different
sizes and types for better generalization.
We train the model using SF graphs of size 10,15,20, and test the
trained model on SF and ER graphs of various sizes up to d=80. As shown in Table \ref{tbl:ensemble}, although the model is only trained on only small graphs, it can work reasonably well even on large graphs of size d=80.

\begin{table}[ht]
\centering
\footnotesize
\caption{\small Ensemble training on SF graphs of size 10,15,20. Testing on ER and SF graphs of sizes up to d=80. \label{tbl:ensemble}}
\begin{tabular}{l|r|crrr|crrr}
\toprule
model       & test\_d       &test\_gtype & prec & recall & shd    &test\_gtype & prec & recall & shd \\
\midrule
DAG-EQ-Ensemble & 10      & SF          & 89.1 & 95.3   & 1.5     & ER          & 72.1 & 66.9   & 5.6  \\
DAG-EQ-Ensemble & 15      & SF          & 88.2 & 95.7   & 2.4  & ER          & 74.3 & 67.5   & 8.1    \\
DAG-EQ-Ensemble & 20      & SF          & 84.8 & 94.7   & 4.2  & ER          & 72.0 & 66.0   & 11.6   \\
\midrule
DAG-EQ-Ensemble & 30      & SF          & 75.1 & 93.3   & 10.9   & ER          & 68.1 & 68.9   & 18.7 \\
DAG-EQ-Ensemble & 50      & SF          & 58.8 & 90.4   & 35.9   & ER          & 56.0 & 73.3   & 41.8  \\
DAG-EQ-Ensemble & 80      & SF          & 43.4 & 88.6   & 100.2    & ER          & 44.5 & 79.1   & 95.3\\

\bottomrule
\end{tabular}
\end{table}

\subsection{Real Data Experiment}

We test our model on a well-known  real benchmark of a protein signaling network~\cite{2005-Science-Sachs-Causal}. The data contains 853 samples, and the ground truth graph has 11 nodes and 17 edges.
We apply the model ensemble trained using d=10,15,20 graphs above on this dataset. Our trained model outputs 10 edges, with SHD=16, comparable with other methods, shown in Table~\ref{tbl:sachs}.

\begin{table}[ht]
  \centering
  \small
  \caption{Performance comparison on real data~\cite{2005-Science-Sachs-Causal}.\label{tbl:sachs}}
  \begin{tabular}{l|r|r|r|r}
    \toprule
    model & predicted edges & correct edges & SHD \\
    \midrule
    DAG-EQ & 10 & 5 & 16 \\
    NOTEARS & 20 & 6 & 19 \\
    RL-BIC2 & 10 & 7 & 11 \\
    CAM & 10 & 6 & 12 \\
    DAG-GNN & 15 & 6 & 16 \\
    \bottomrule
  \end{tabular}
\end{table}

\section{Conclusion and Future Directions}

We proposed the first supervised approach for full DAG
discovery, and proposed to use equivariant models which fit well for this learning problem. We evaluated the proposed DAG-EQ approach  extensively, and demonstrated the advantage of our whole DAG formulation over pairwise one, the effectiveness of equivariant models over baselines, and promising results in comparison  with state-of-the-art optimization-based methods.

We hope the formulation of DAG discovery into supervised learning will inspire many future works. 
On model design, we are interested in exploring other network structures (such as graph neural networks) and enforcing acyclic constraint (e.g. the equality constraint in NOTEARS). On problem setting, it is interesting to extend the model to non-linear causal models and interventional data. It is also interesting to explore more flexible input distribution featurizations that can be learned and adapted automatically for different datasets. 
\bibliography{paper}
\bibliographystyle{apalike}


\end{document}